\documentclass{article} 
\usepackage{iclr2021_conference,times}

\usepackage{amsmath,amsfonts,bm}

\def\eqref#1{equation~\ref{#1}}

\def\1{\bm{1}}

\def\rvx{{\mathbf{x}}}

\def\rvz{{\mathbf{z}}}

\def\vb{{\bm{b}}}

\def\vu{{\bm{u}}}
\def\vv{{\bm{v}}}

\def\vx{{\bm{x}}}
\def\vy{{\bm{y}}}

\def\mA{{\bm{A}}}

\def\mI{{\bm{I}}}

\def\mT{{\bm{T}}}

\def\mX{{\bm{X}}}
\def\mY{{\bm{Y}}}

\DeclareMathAlphabet{\mathsfit}{\encodingdefault}{\sfdefault}{m}{sl}
\SetMathAlphabet{\mathsfit}{bold}{\encodingdefault}{\sfdefault}{bx}{n}
\newcommand{\tens}[1]{\bm{\mathsfit{#1}}}
\def\tA{{\tens{A}}}
\def\tB{{\tens{B}}}

\def\emA{{A}}

\usepackage{hyperref}
\usepackage{url}
\usepackage{graphicx}
\usepackage{wrapfig}
\usepackage{amssymb}
\usepackage[ruled, vlined]{algorithm2e}
\usepackage{setspace}
\usepackage{caption}
\usepackage{subcaption}

\title{Training Invertible Linear Layers through Rank-One Perturbations}

\author{
Andreas Krämer, Jonas Köhler, Frank Noé\thanks{also at Rice University, Dept. of Chemistry, Houston, TX 77005, USA, and FU Berlin, Dept. of Physics} \\
Department of Mathematics and Computer Science\\
Freie Universität Berlin\\
\texttt{\{andreas.kraemer, jonas.koehler, frank.noe\}@fu-berlin.de} \\
}

\iclrfinalcopy 
\iclrpreprintcopy 
\begin{document}

\maketitle

\begin{abstract}
Many types of neural network layers rely on matrix properties
such as invertibility or orthogonality.
Retaining such properties during optimization with gradient-based 
stochastic optimizers is a challenging task, which is usually addressed
by either reparameterization of the affected parameters or 
by directly optimizing on the manifold.
This work presents a novel approach for training invertible linear layers. In lieu of directly optimizing
the network parameters, we train rank-one perturbations and add them to the actual weight matrices infrequently. This \textit{P$^{4}$Inv update} allows keeping track of inverses and determinants without ever explicitly computing them. We show how such invertible blocks improve the mixing and thus the mode separation of the resulting normalizing flows. Furthermore, we outline how the \textit{P$^4$} concept can be utilized to retain properties other than invertibility.
\end{abstract}

\section{Introduction}

\begin{wrapfigure}{R}{0.5\textwidth}
\begin{center}
\includegraphics[width=\linewidth]{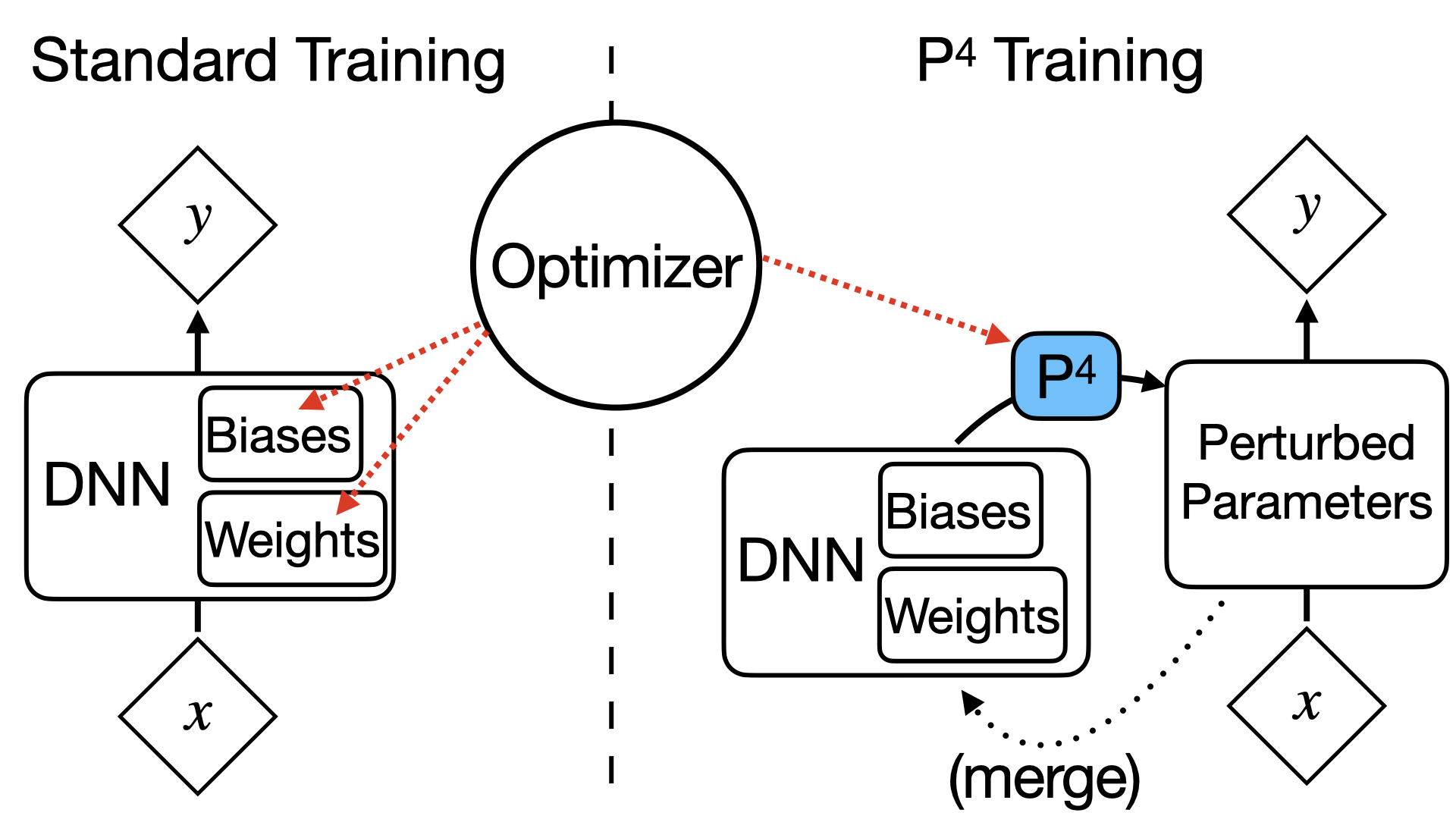}
\end{center}
\caption{Training of deep neural networks (DNN). Standard DNN transform inputs $x$ into outputs $y$ through activation functions and linear layers, which are tuned by an optimizer. In contrast, P$^4$ training operates on perturbations to the parameters. Those are defined to retain certain network properties (here: invertibility as well as tractable inversion and determinant computation). The perturbed parameters are merged in regular intervals.
\label{fig:p4}
}
\end{wrapfigure}

Many deep learning applications depend critically on the neural network parameters having a certain mathematical structure. As an important example, reversible generative models rely on invertibility and, in the case of normalizing flows, efficient inversion and computation of the Jacobian determinant \citep{papamakarios2019normalizing}.

Preserving parameter properties during training can be challenging and many approaches are currently in use. The most basic way of incorporating constraints is by network design. Many examples could be listed, like defining convolutional layers to obtain equivariances, constraining network outputs to certain intervals through bounded activation functions, Householder flows \citep{tomczak2016improving} to enforce layer-wise orthogonality, or coupling layers \citep{dinh2014nice, dinh2016density} that enforce tractable inversion through their two-channel structure.
A second approach concerns the optimizers used for training. Optimization routines have been tailored for example to maintain Lipschitz bounds \citep{yoshida2017spectral} or efficiently optimize orthogonal linear layers \citep{choromanski2020stochastic}.

The present work introduces a novel algorithmic concept for training invertible linear layers and facilitate tractable inversion and determinant computation, see Figure \ref{fig:p4}. In lieu of directly changing the network parameters, the optimizer operates on perturbations to these parameters. The actual network parameters are frozen, while a parameterized perturbation (a rank-one update to the frozen parameters) serves as a proxy for optimization. Inputs are passed through the perturbed network during training. In regular intervals, the perturbed parameters are merged into the actual network and the perturbation is reset to the identity. This stepwise optimization approach will be referred to as property-preserving parameter perturbation, or \textit{P$^4$ update}. A similar concept was introduced recently by \cite{lezcano2019trivializations}, who used dynamic trivializations for optimization on manifolds.

In this work, we use P$^4$ training to optimize invertible linear layers while keeping track of their inverses and determinants using rank-one updates. Previous work (see Section \ref{sec:related}) has mostly focused on optimizing orthogonal matrices, which can be trivially inverted and have unity determinant. Only most recently, \cite{gresele2020relative} presented a first method to optimize general invertible matrices implicitly using relative gradients, thereby providing greater flexibility and expressivity. While their scheme implicitly tracks the weight matrices' determinants, it does not facilitate cheap inversion. In contrast, the present \textit{P$^{4}$Inv layers} are inverted at the cost of roughly three matrix-vector multiplications. 

P$^{4}$Inv layers can approximate arbitrary invertible matrices $\mA \in \mathrm{GL}(n)$. Interestingly, our stepwise perturbation even allows sign changes in the determinants and recovers the correct inverse after emerging from the ill-conditioned regime. Furthermore, it avoids any explicit computations of inverses or determinants. All operations occurring in optimization steps have complexity of at most $\mathcal{O}(n^2).$ To our knowledge, the present method is the first to feature these desirable properties.

We show how P$^4$Inv blocks can be utilized in normalizing flows by combining them with nonlinear, bijective activation functions and with coupling layers. The resulting neural networks are validated for density estimation and as deep generative models. Finally, we outline potential applications of P$^4$ training to network properties other than invertibility.

\section{Background and Related Work}
\label{sec:related}

\subsection{Rank-one Perturbation}
The P$^4$Inv layers are based on rank-one updates, which are defined as transformations $\mA\mapsto \mA+\vu\vv^{T}$ with
$\vu,\vv \in \mathbb{R}^{n}$. If $\mA \in \mathrm{GL}(n)$ and $1+\vv^{T}\mA^{-1}\vu\neq0$,
the updated matrix is also invertible and its inverse can be computed
by the Sherman-Morrison formula
\begin{equation}
    (\mA+\vu \vv^{T})^{-1} 
    =
    \mA^{-1}-\frac{1}{1+\vv^{T}\mA^{-1}\vu}\mA^{-1}\vu \vv^{T} \mA^{-1}.
    \label{eq:rankone}
\end{equation}
Furthermore, the determinant is given by the matrix determinant lemma
\begin{equation}
\det(\mA+\vu\vv^{T})=(1+\vv^{T}\mA^{-1}\vu)\det(\mA).\label{eq:rankone_det}
\end{equation}
Both these equations are widely used in numerical mathematics, since they sidestep the $\mathcal{O}(n^{3})$ cost and poor parallelization of both matrix inversion and determinant computation. The present work leverages these perturbation formulas to keep track of the inverses and determinants of weight matrices during training of invertible neural networks.

\subsection{Existing Approaches for Training Invertible Linear Layers}
Maintaining invertibility of linear layers has been studied in the context of convolution operators
\citep{kingma2018glow, karami2019invertible, hoogeboom2019emerging, hoogeboom2020convolution} and using Sylvester's theorem \citep{berg2018sylvester}. Those approaches often involve decompositions that include triangular matrices (\cite{papamakarios2019normalizing}). While inverting triangular matrices has quadratic computational complexity, it is inherently sequential and thus fairly inefficient on parallel computers (see Section \ref{sec:cost}).  More closely related to our work, \cite{gresele2020relative} introduced a relative gradient optimization scheme for invertible matrices. In contrast to this related work, our method facilitates a cheap inverse pass and allows sign changes in the determinant. On the contrary, their method operates in a higher-dimensional search space, which might speed up the optimization in tasks that do not involve inversion during training.

\subsection{Normalizing Flows}
Cheap inversion and determinant computation are specifically important in the context of normalizing flows, see Appendix \ref{sec:nf}.
They were introduced in  \cite{tabak2010density, tabak2013family} and are commonly used, either in variational inference \citep{rezende2015variational, tomczak2016improving, louizos2017multiplicative, berg2018sylvester} or for approximate sampling from distributions given by an energy function \citep{oord2017parallel,muller2018neural, noe2019boltzmann, kohler2020equivariant}.
The most important normalizing flow architectures are coupling layers \citep{dinh2014nice, dinh2016density, kingma2018glow, muller2018neural}, which are a subclass of autoregressive flows \citep{germain2015made, papamakarios2017masked, huang2018neural, de2019block}, and (2)
residual flows \citep{chen2018neural, zhang2018monge, grathwohl2018ffjord, behrmann2018invertible, chen2019residual}.
A comprehensive survey can be found in \cite{papamakarios2019normalizing}.

\subsection{Optimization under Constraints and Dynamic Trivializations}
Constrained matrices can be optimized using Riemannian gradient descent on the manifold (\cite{Absil2009}).
A reparameterization trick for general Lie groups has been introduced in
\citet{falorsi2019reparameterizing}. 
For the unitary / orthogonal group there are multiple more specialized approaches, including using the 
Cayley transform
\citep{helfrich2018orthogonal},
Householder Reflections
\citep{mhammedi2017efficient, meng2020gaussianization, tomczak2016improving},
Givens rotations
\citep{Shalit2014, pevny2020sum} or the 
exponential map
\citep{lezcano2019cheap, golinski2019improving}.

\cite{lezcano2019trivializations} introduced the concept of dynamic trivializations. This method performs training on manifolds by combining ideas from Riemannian gradient descent and trivializations (parameterizations of the manifold via an unconstrained space). Dynamic trivializations were derived in the general settings of Riemannian exponential maps and Lie groups. Convergence results were recently proven in follow-up work (\cite{lezcanocasado2020curvaturedependant}). P$^4$ training resembles dynamic trivializations in that both perform a number of iteration steps in a fixed basis and infrequently lift the optimization problem to a new basis. 
In contrast, the rank-one updates do not strictly parameterize $\mathrm{GL}(n)$ but instead can access all of $\mathbb{R}^{n\times n}.$ This introduces the need for numerical stabilization, but enables efficient computation of the inverse and determinant through \eqref{eq:rankone} and \eqref{eq:rankone_det}, which is the method's unique and most important aspect.

\section{P$^4$ Updates: Preserving Properties through Perturbations}

\subsection{General Concept}
\label{sec:general}
A deep neural network is a parameterized function $M_\tA:\mathbb{R}^n\rightarrow \mathbb{R}^m$ with a high-dimensional parameter tensor $\tA.$
Now, let $\mathbb{S}$ define the subset of feasible
parameter tensors so that the network satisfies a certain desirable property. 
In many situations, generating elements of $\mathbb{S}$ from scratch is much harder than transforming any $\tA \in \mathbb{S}$ into other
elements $\tA^\prime \in \mathbb{S}$, i.e. to move within $\mathbb{S}$.

The efficiency of perturbative updates can be leveraged as an incremental approach to retain certain desirable
properties of the network parameters during training. Rather than
optimizing the parameter tensors directly, we instead use a
transformation 
$
R_\tB:\mathbb{S}\rightarrow \mathbb{S},
$
which we call a \textit{property-preserving parameter perturbation}
(P$^4$). A P$^4$ transforms a given parameter tensor $\tA \in \mathbb{S}$
into another tensor with the desired property $\tA^\prime \in \mathbb{S}.$ The P$^4$ itself is also parameterized, by a tensor $\tB.$ We demand that the identity $id_{\mathbb{S}}: \tA \mapsto \tA$ be included in the set of these transformations, i.e. there exists
a $\tB_0$ such that $R_{\tB_0}=id_{\mathbb{S}}$.

During training, the network is evaluated using the perturbed parameters
$\tilde{\tA}=R_{\tB}(\tA).$ The parameter tensor of the perturbation, $\tB,$ is trainable via gradient-based stochastic optimizers, while the actual parameters $\tA$ are frozen.
In regular intervals, every $N$ iterations of the optimizer, the optimized parameters of the P$^4$, $\tB$, are merged into $\tA$ as follows:
\begin{align}
    \tA_{\mathrm{new}} &\leftarrow R_\tB (\tA), \label{eq:merge1} \\
    \tB_{\mathrm{new}} &\leftarrow \tB_0. \label{eq:merge2}
\end{align}
This update does not modify the effective (perturbed) parameters of the network $\tilde{\tA}$, since
$$
\tilde{\tA}_{\mathrm{new}} = R_{\tB_\mathrm{new}}(\tA_\mathrm{new}) = R_{\tB_0}(R_{\tB}(\tA)) = R_{\tB}(\tA) = \tilde{\tA}.
$$
Hence, this procedure enables a steady, iterative transformation
of the effective network parameters and stochastic gradient descent
methods can be used for training without major modifications.
Furthermore, given a reasonable P$^4$, the iterative update of $\tA$
can produce increasingly non-trivial transformations, thereby enabling high expressivity of the resulting neural networks. This concept is summarized in Algorithm \ref{alg:p4}. Further extensions to stabilize the merging step will be exemplified in Section \ref{sec:stabilization}.

\begin{algorithm}[H]
\SetAlgoLined
\SetKwInOut{Input}{Input}
\SetKwInOut{Output}{Output}
\Input{Model $M$, training data, loss function $J$, number of optimization steps $N_\mathrm{steps}$, merge interval $N$, perturbation $R$, optimizer OPT}
 initialize $\tA \in \mathbb{S}$; \\
 initialize $\tB$ := $\tB_0$; \\
 \For{$i := 1 \dots N_\mathrm{steps}$}{
 $\mX, \mY_0$ := $i$-th batch from training data; \\
 $\tilde{\tA}$ := $R_{\tB}(\tA)$ \tcp*[r]{perturb parameters}
 $\mY$ := $M_{\tilde{\tA}}(\mX)$ \tcp*[r]{evaluate perturbed model}
 j := $J(\mY, \mY_0)$ \tcp*[r]{evaluate loss function}
 gradient := $\partial j / \partial \tB$ \tcp*[r]{backpropagation}
 $\tB$ := OPT$(\tB, \mathrm{gradient})$ \tcp*[r]{optimization step}
  \If{$i \mod N = 0$}{
   $\tA$ := $R_\tB (\tA)$  \tcp*[r]{merging step: update frozen parameters}
   $\tB$ := $\tB_0$ \tcp*[r]{merging step: reset perturbation}
   }
 }
\caption{P$^4$ Training \label{alg:p4}}
\end{algorithm}

\subsection{P$^4$Inv: Invertible Layers via Rank-One Updates}
\begin{algorithm}[H]
\SetAlgoLined
\SetKwInOut{Input}{Input}
\SetKwInOut{Output}{Output}
\Input{Matrix $\mA$, Inverse $\mA_\mathrm{inv}$, Determinant $d$}
   det\_factor := $(1+\vv^{T}\mA_\mathrm{inv}\vu)$ \\
   new\_det := det\_factor $\cdot$ $d$; \\
   \If{$\ln | \mathrm{det}\_\mathrm{factor}|$ and $\ln | \mathrm{new}\_\mathrm{det}|$ are sane}{
     \tcc{update frozen parameters (\eqref{eq:merge1})}
       $d$ := new\_det; \\
       $\mA$ := $R_{\vu, \vv} (\mA)$; \\
       $\mA_\mathrm{inv}$ := $\mA_\mathrm{inv}-\frac{1}{1+\vv^{T}\mA_\mathrm{inv}\vu}\mA_\mathrm{inv}\vu \vv^{T} \mA_\mathrm{inv}$; \\
      \tcc{reset perturbation (\eqref{eq:merge2})}
       $\vu$ := $0$; \\
       $\vv$ := $\mathcal{N}(0,\mI_n)$ \tcp*[r]{random reinitialization}
   }
\caption{P$^4$Inv Merging Step \label{alg:p4inv}}
\end{algorithm}

The P$^4$ algorithm can in principle be applied to properties concerning either individual blocks or the whole network. Here we train individual invertible linear layers via rank-one perturbations. Each of these P$^4$Inv layers is an affine transformation $\mA \vx + \vb.$ In this context, the weight matrix $\mA$ is handled by the P$^4$ update and the bias $\vb$ is optimized without perturbation. Without loss of generality, we present the method for layers $\mA \vx.$

We define $\mathbb{S}$ as the set of invertible
matrices, for which we know the inverse and determinant. Then the rank-one update 
\begin{equation}
R_{\vu,\vv}(\mA)=\mA+\vu \vv^{T}
\end{equation}
with $\tB = (\vu,\vv)\in\mathbb{R}^{2n}$ is a P$^4$ on $\mathbb{S}$ due to equations \ref{eq:rankone} and \ref{eq:rankone_det}, which also define the inverse pass and determinant computation of the perturbed layer, see Appendix \ref{sec:implementation} for details. The perturbation can be reset by setting $\vu$, $\vv$, or both to zero. In subsequent parameter studies, a favorable training efficiency was obtained by setting $\vu$ to zero and reinitializing $\vv$ from Gaussian noise. (Using a unity standard deviation for the reinitialization ensures that gradient-based updates to $u$ are on the same order of magnitude as updates to a standard linear layer so that learning rates are transferable.) The inverse matrix $\mA_\mathrm{inv}$ and determinant $d$ are stored in the P$^4$ layer alongside $\mA$ and updated according to the merging step in Algorithm \ref{alg:p4inv}. Merges are skipped whenever the determinant update signals ill conditioning of the inversion. This is further explained in the following subsection.

\subsection{Numerical Stabilization}
\label{sec:stabilization}
The update to the inverse and determinant can become ill-conditioned
if the denominator in equation \ref{eq:rankone} is close to zero.
Thankfully, the determinant lemma from \eqref{eq:rankone_det} provides an indicator for ill-conditioned updates (if absolute determinants become very small or very large).
This indicator in combination with the stepwise merging approach can be used to tame potential numerical issues. Concretely, the following additional strategies are applied to ensure stable optimizations.

\begin{itemize}
\item \emph{Skip Merges:} Merges are skipped whenever the determinant update falls out of predefined bounds, see Appendix \ref{sec:sanity} for details. This allows the optimization to continue without propagating numerical errors into the actual weight matrix $\mA$. Note that numerical errors in the perturbed parameters $\tilde{\mA}$ are instantaneous and vanish when the optimization leaves the ill-conditioned regime. As shown in our experiments in Section \ref{sec:linear}, merging steps that occur relatively infrequently without drastically hurting the efficiency of the optimization. 

\item \emph{Penalization:} The objective function can be augmented by a penalty function $g(\vu,\vv)$ in order to prevent entering the ill-conditioned regime $\left\{ (\vu, \vv) : \det\left(R_{\vu,\vv}(\mA)\right)=0\right\},$ see Appendix \ref{sec:sanity}.

\item \emph{Iterative Inversion:} In order to maintain a small error of the inverse throughout training, the inverse is corrected after every $N_{\mathrm{correct}}$-th merging step by one iteration of an iterative matrix inversion \citep{soleymani2014fast}. This operation 
is $O(n^3)$
yet is highly parallel.
\end{itemize}

\subsection{Use in Invertible Networks}
\label{sec:usage}

Our invertible linear layers can be employed in normalizing flows (Appendix \ref{sec:nf}) thanks to having access to the determinant at each update step. We tested them in two different application scenarios:

\paragraph{P$^4$Inv Swaps}

In a first attempt, we integrate P$^4$Inv layers with \textit{RealNVP} coupling layers by substituting the simple coordinate swaps with general linear layers (see Figure \ref{fig:bgarchitecture} in Appendix \ref{sec:trainbg}). Fixed coordinate swaps span a tiny subset of $\mathrm{O}(n)$. In contrast, P$^4$Inv can express all of $\mathrm{GL}(n)$. We thus expect more expressivity with the help of better mixing. The parameter matrix $\mA$ is initialized with a permutation matrix. Note that the P$^4$ training is applied exclusively to the P$^4$Inv layers rather than all parameters.

\paragraph{Nonlinear invertible layer}
In a second attempt, we follow the approach of \cite{gresele2020relative} and stack P$^4$Inv layers with intermediate bijective nonlinear activation functions. Here we use the elementwise Bent identity
\begin{equation*}
    B(x) = \frac{\sqrt{x^2+1} - 1}{2} + x.
\end{equation*}
In contrast to more standard activation functions like sigmoids or ReLU variants, the Bent identity is an $\mathbb{R}$-diffeomorphism. It thereby provides smooth gradients and is invertible over all of $\mathbb{R}.$

\section{Experiments}
P$^4$Inv updates are demonstrated in three steps. After a runtime comparison, single P$^4$Inv layers are first fit to linear problems to explore their general capabilities and limitations. Second, to show their performance in deep architectures, P$^4$Inv blocks are used in combination with the Bent identity to perform density estimation of common two-dimensional distributions. Third, to study the generative performance of normalizing flows that use P$^4$Inv blocks, we train a RealNVP normalizing flow with P$^4$ swaps as a Boltzmann generator \citep{noe2019boltzmann}. One important feature of this test problem is the availability of a ground truth energy function that is highly sensitive to any numerical problems in the network inversion.

\subsection{Computational Cost}
\label{sec:cost}

\begin{figure}[tbhp]
\begin{center}
\includegraphics[width=0.9\linewidth]{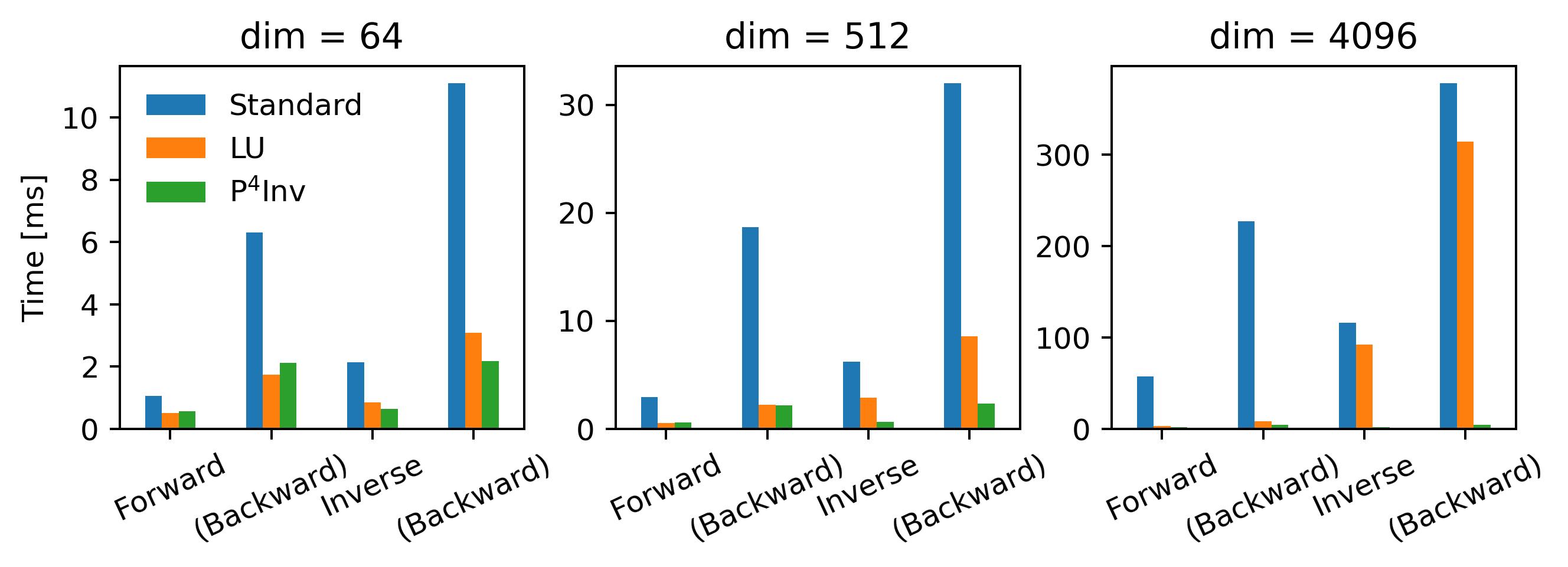}
\end{center}
\caption{
Wall-clock times of forward and backward pass of linear normalizing flows including determinant computation. Three methods are compared: (a) standard linear layers (``standard''), where the inverses and determinants are computed through PyTorch's \emph{inverse} and \emph{det} functions; (b) LU decompositions (``LU''), where the determinants are products over the diagonal entries and the matrices are inverted through \emph{triangular\_solve}; (c) P$^4$Inv updates that keep track of inverses and determinants through rank-one updates. Timings are compared for square matrices of dimension 64, 512, and 4096.
\label{fig:timings}
}
\end{figure}

P$^4$Inv training facilitates cheap inversion and determinant computation. To demonstrate those benefits, the computational cost of computing the forward and inverse KL divergence in a normalizing flow framework was compared with standard linear layers and an LU decomposition. Importantly, the KL divergence includes a network pass and the Jacobian determinant.

Figure \ref{fig:timings} shows the wall-clock times per evaluation on an NVIDIA GeForce GTX 1080 card with a batch size of 32. As the matrix dimension grows, standard linear layers become increasingly infeasible due to the $O(n^3)$ cost of both determinant computation and inversion. The LU decomposition is fast for forward evaluation since the determinant is just the product of diagonal entries. However, the inversion does not parallelize well so that inverse pass of a 4096-dimensional matrix was almost as slow as a naive inversion. Note that this poor performance transfers to other decompositions involving triangular matrices, such as the Cholesky decomposition. In contrast, the P$^4$Inv layers performed well for both the forward and inverse evaluation. Due to their $O(n^2)$ scaling, they outperformed the two other methods by two orders of magnitude on the 4096-dimensional problem.

This comparison shows that P$^4$Inv layers are especially useful in the context of normalizing flows whose forward and inverse have to be computed during training. This situation occurs when flows are trained through a combination of density estimation and sampling. 

\subsection{Linear Problems}
\label{sec:linear}
Fitting linear layers to linear training data is trivial in principle using basic linear algebra methods. However, the optimization with stochastic gradient descent at a small learning rate will help illuminate some important capabilities and limitations of P$^4$Inv layers. It will also help answer the open question if gradient-based optimization of an invertible matrix $\mA$ allows crossing the ill-conditioned regime $\{\mA \in \mathbb{R}^{n\times n}: \det \mA = 0 \}.$ Furthermore, the training efficiency of perturbation updates can be directly compared to arbitrary linear layers that are optimized without perturbations.

Specifically, each target problem is defined by a square matrix $\mT.$ The training data is generated by sampling random vectors $\vx$ and computing targets $\vy=\mT \vx.$ Linear layers are then initialized as the identity matrix $\mA:=\mI$ and the loss function $J(\mA) = \| \mA \vx - \vy \|^2$ is minimized in three ways:
\begin{samepage}
\begin{enumerate}
    \item by directly updating the matrix elements (standard training of  linear layers),
    \item through P$^4$Inv updates, and
    \item through the inverses of P$^4$Inv updates, i.e., by training $\mA$ through the updates in \eqref{eq:rankone}.
\end{enumerate}
\end{samepage}

\begin{figure}[tbhp]
\centering
\includegraphics[width=7.0cm]{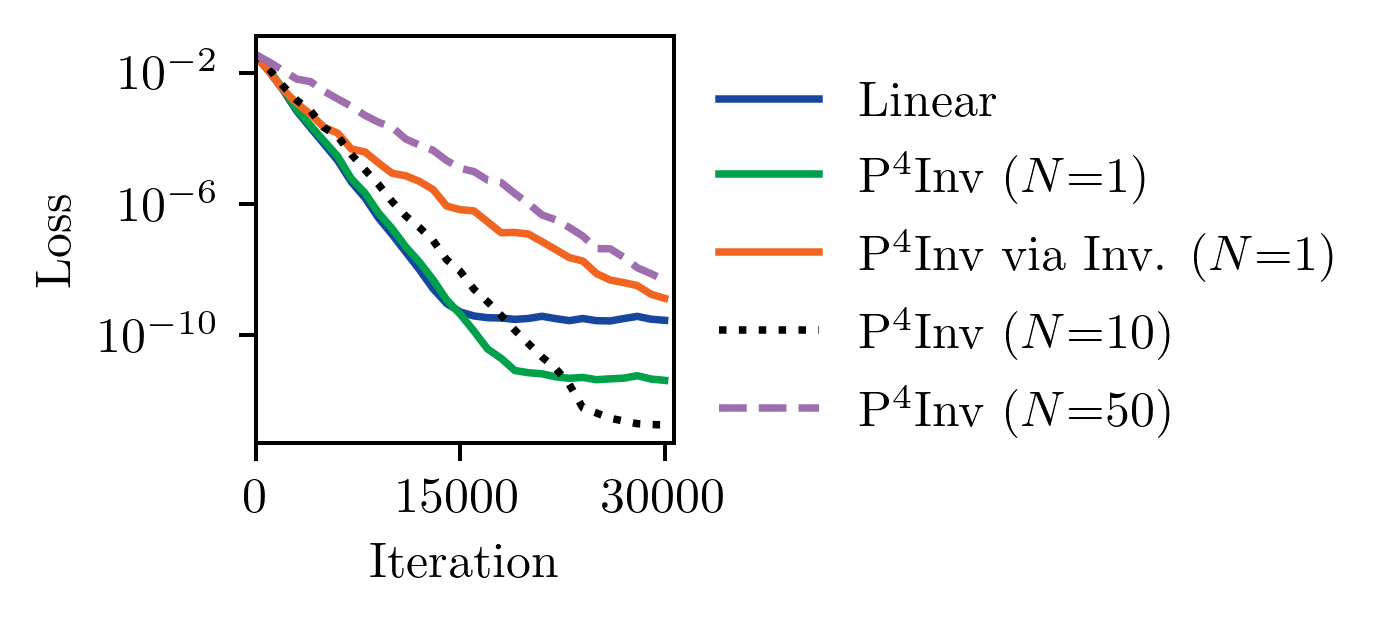}
\includegraphics[width=6.8cm]{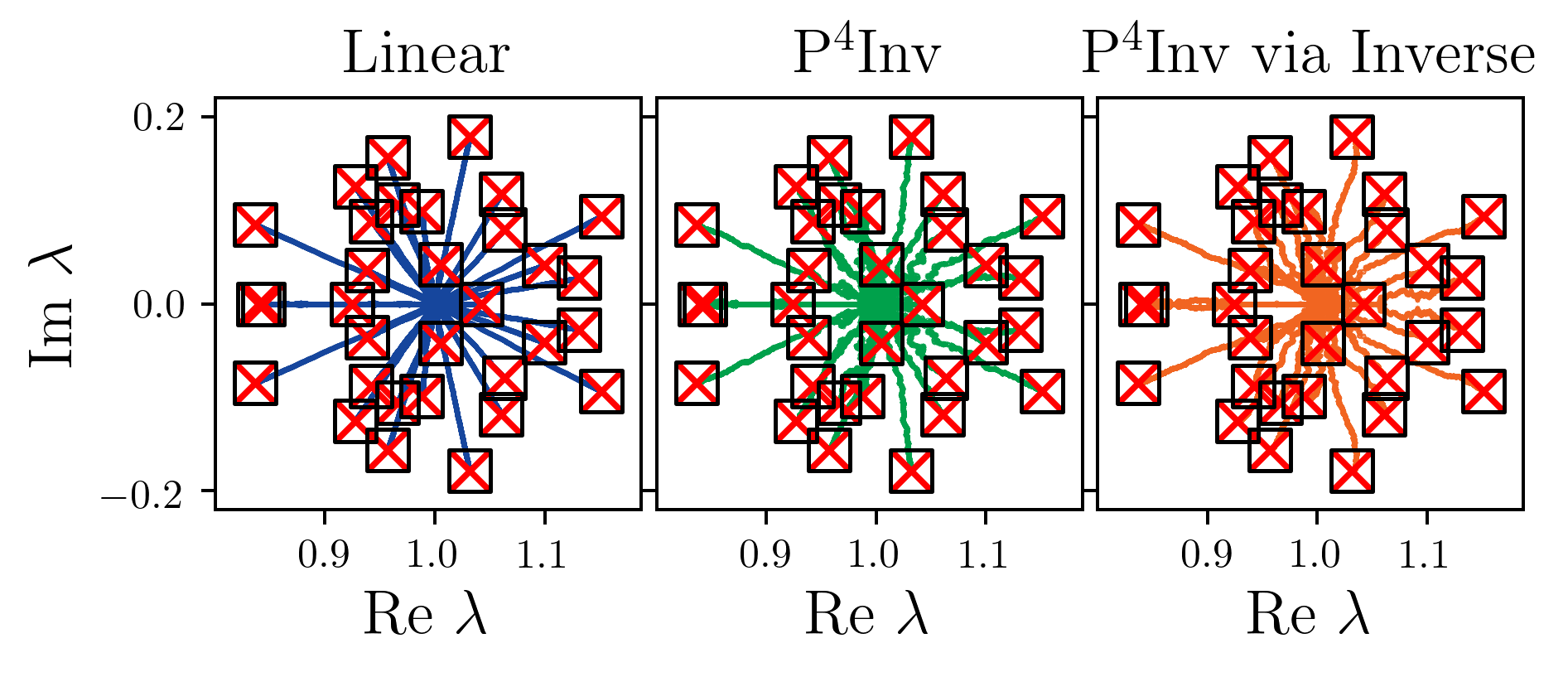}
\caption{
Training towards a 32-dimensional positive definite target matrix $\mT$. Left: Losses during training. Right: Eigenvalues during training. Final eigenvalues are shown as red crosses. Eigenvalues of the target matrix are shown as black squares. 
\label{fig:posdef}
}
\end{figure}

The first linear problem is a 32-dimensional positive definite matrix with eigenvalues close to 1. Figure \ref{fig:posdef} shows the evolution of eigenvalues and losses during training. All three methods of optimization successfully recovered the target matrix. While training P$^4$Inv via the inverse led to slower convergence, the forward training of P$^4$Inv converged in the same number of iterations as an unconstrained linear layer for a merge interval $N=1$. Increasing the merge interval to $N=10$ only affected the convergence minimally. Even for $N=50,$ the optimizer took only twice as many iterations as for an unconstrained linear layer. 

\begin{figure}[tbhp]
\begin{center}
\includegraphics[width=7.0cm]{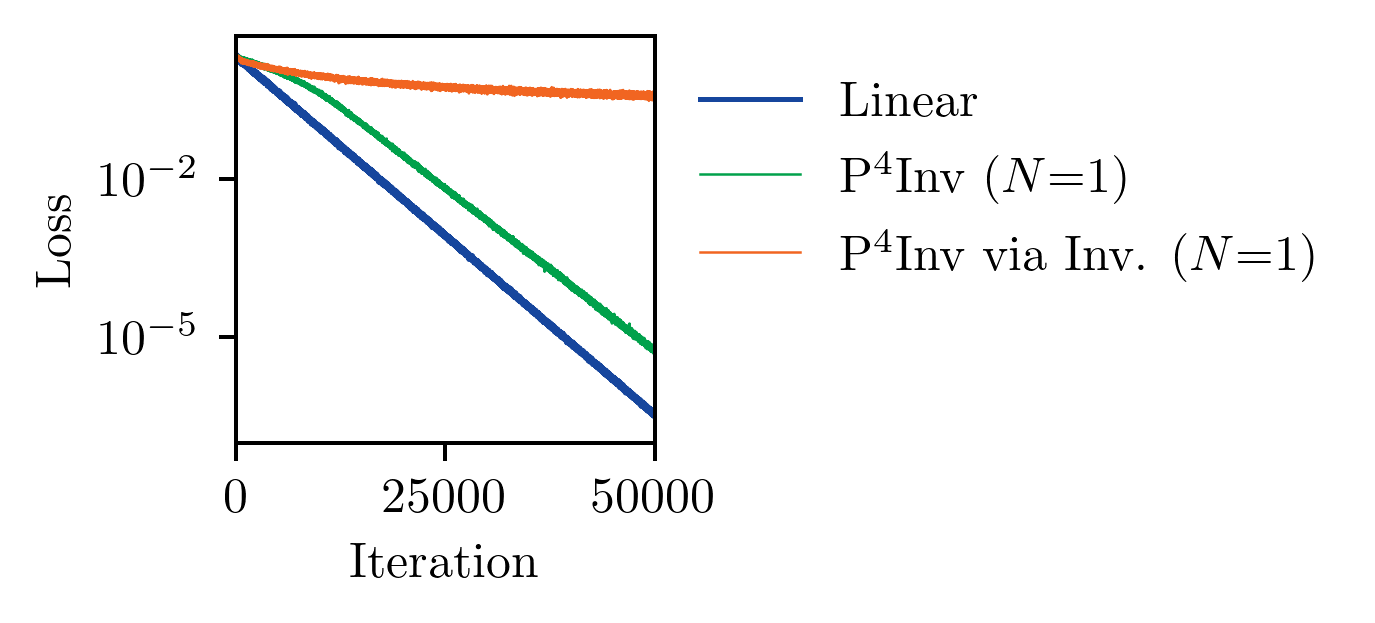}
\includegraphics[width=6.8cm]{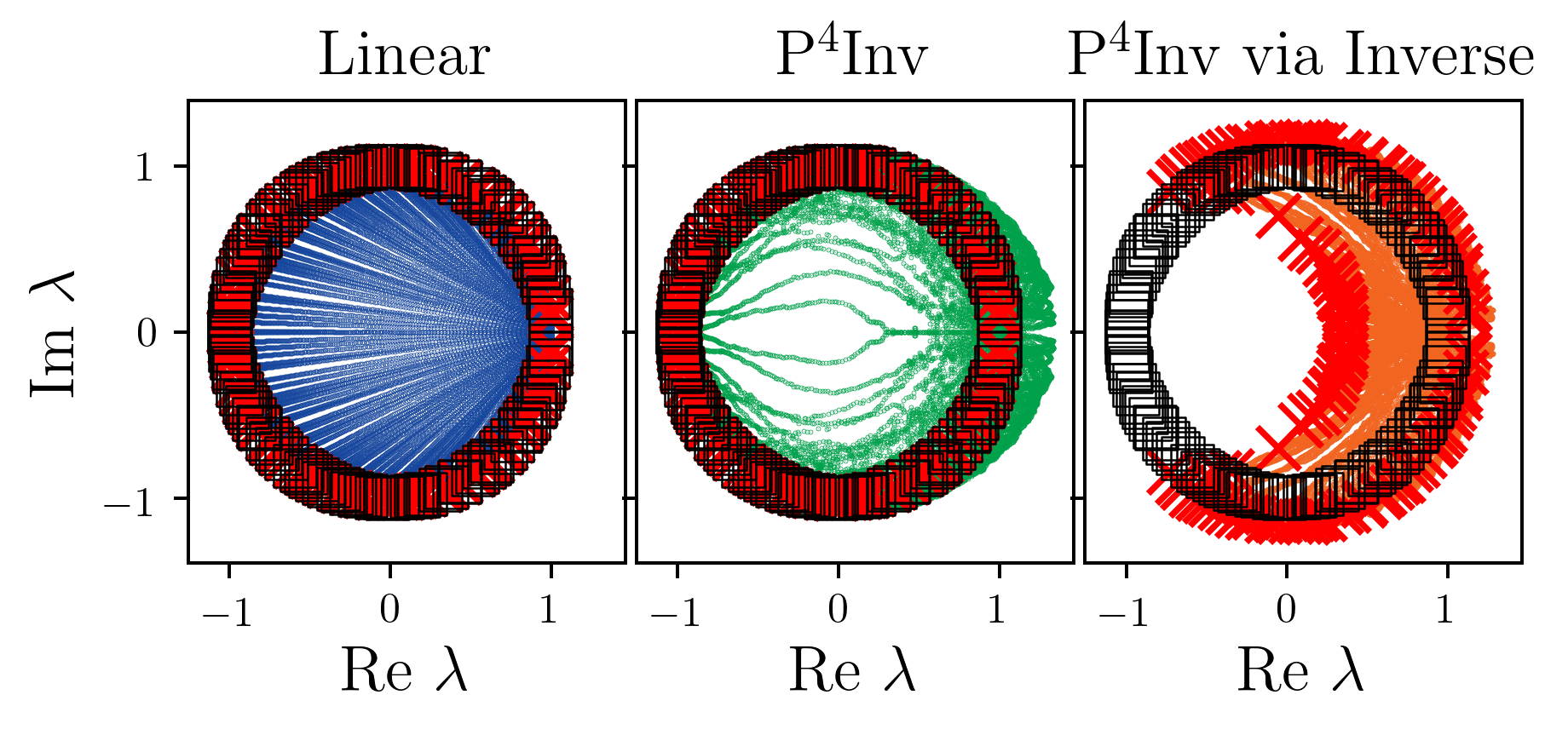}
\end{center}
\caption{
Training towards an orthogonal target matrix $\mT \in \mathrm{SO}(128)$. Left: Losses during training. Right: Eigenvalues during training. Final eigenvalues are shown as red crosses. Eigenvalues of the target matrix are shown as black squares. 
\label{fig:orthogonal}
}
\end{figure}

The second target matrix was a 128-dimensional special orthogonal matrix. As shown in Figure \ref{fig:orthogonal}, the direct optimization converged to the target matrix in a linear fashion. In contrast, the matrices generated by the P$^4$Inv update avoided the region around the origin. This detour led to a slower convergence in the initial phase of the optimization. Notably, the inverse stayed accurate up to 5 decimals throughout training. Training an inverse P$^4$Inv was not successful for this example. This shows that the inverse P$^4$Inv update can easily get stuck in local minima. This is not surprising as the elements of the inverse (\eqref{eq:rankone}) are parameterized by $\mathbb{R}^{2n}$-dimensional rational quadratic functions. When training against linear training data with a unique global optimum, the multimodality can prevent convergence. When training against more complex target data, the premature convergence was mitigated, see Appendix \ref{sec:train2D}. However, this result suggests that the efficiency of the optimization may be perturbed by very complex nonlinear parameterizations.

The final target matrix was $\mT = -\mI_{101},$ a matrix with determinant -1. In order to iteratively converge to the target matrix, the set of singular matrices has to be crossed. As expected, using a nonzero penalty parameter prevented the P$^4$Inv update from converging to the target. However, when no penalty was applied, the matrix converged almost as fast as the usual linear training, see Figure \ref{fig:negeye}. When the determinant approached zero, inversion became ill-conditioned and residues increased. However, after reaching the other side, the inverse was quickly recovered up to 5 decimal digits. Notably, the determinant also converged to the correct value despite never being explicitly corrected. 

The favorable training efficiency encountered in those linear problems is surprising given the considerably reduced search space dimension. In fact, a subsequent rank-one parameterization of an MNIST classification task suggests that applications in nonlinear settings also converge as fast as standard MLPs in the initial phase, but slow down when approaching the optimum, see Appendix \ref{sec:mnist}.

\subsection{2D Distributions}
The next step was to assess the effectiveness of P$^4$Inv layers in deep networks. This was particularly important to rule out a potentially harmful accumulation of rounding errors. Density estimation of common 2D toy distributions was performed by stacking P$^4$Inv layers with Bent identities and their inverses. For comparison, an RNVP flow was constructed with the same number of tunable parameters as the P$^4$Inv flow, see Appendix \ref{sec:train2D} for details.

\begin{figure}[tbhp]
\centering
\begin{minipage}[t]{.4\textwidth}
    \centering
    \includegraphics[width=\linewidth]{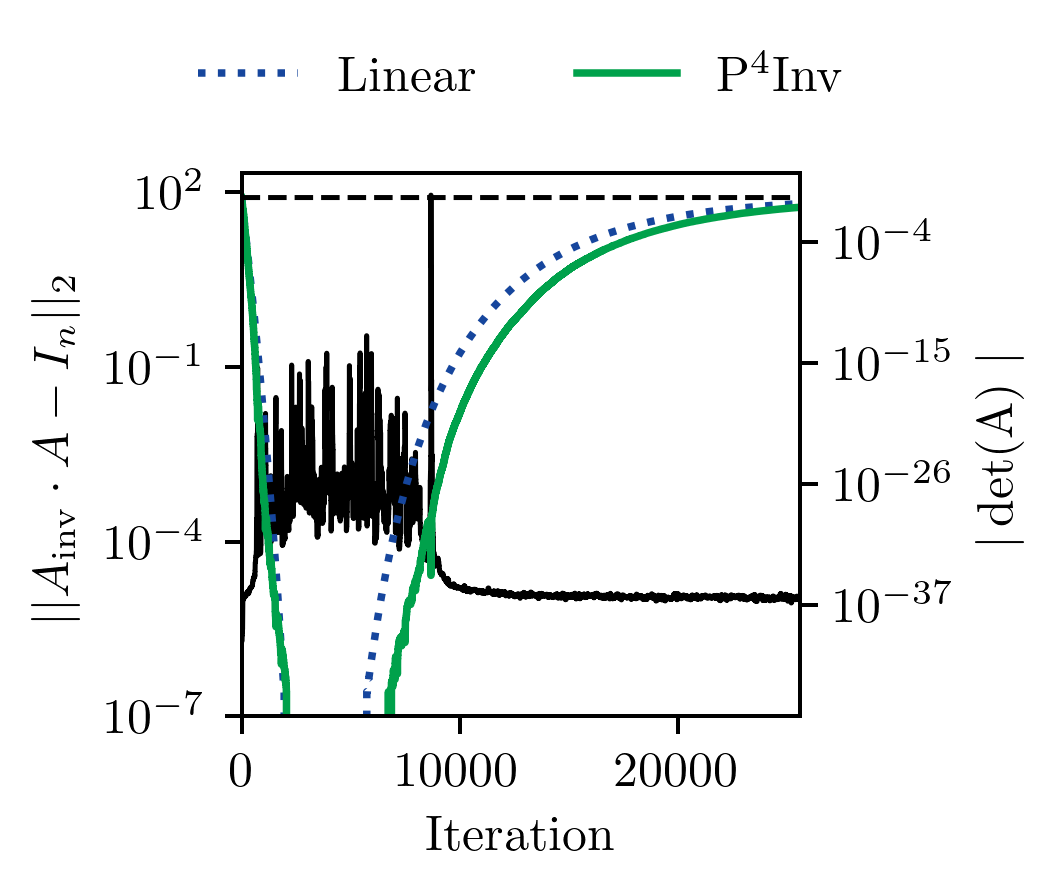}
    \captionof{figure}{
    Training towards the matrix $\mT=-\mI_{101}$ using no penalty. Residue of inversion (black line) and absolute determinants of the standard linear and P$^4$Inv layer. Both converge to the target in a similar number iterations (dashed line).
    \label{fig:negeye}
    }
\end{minipage}%
\begin{minipage}{.6\textwidth}
  \centering
    \includegraphics[width=\linewidth]{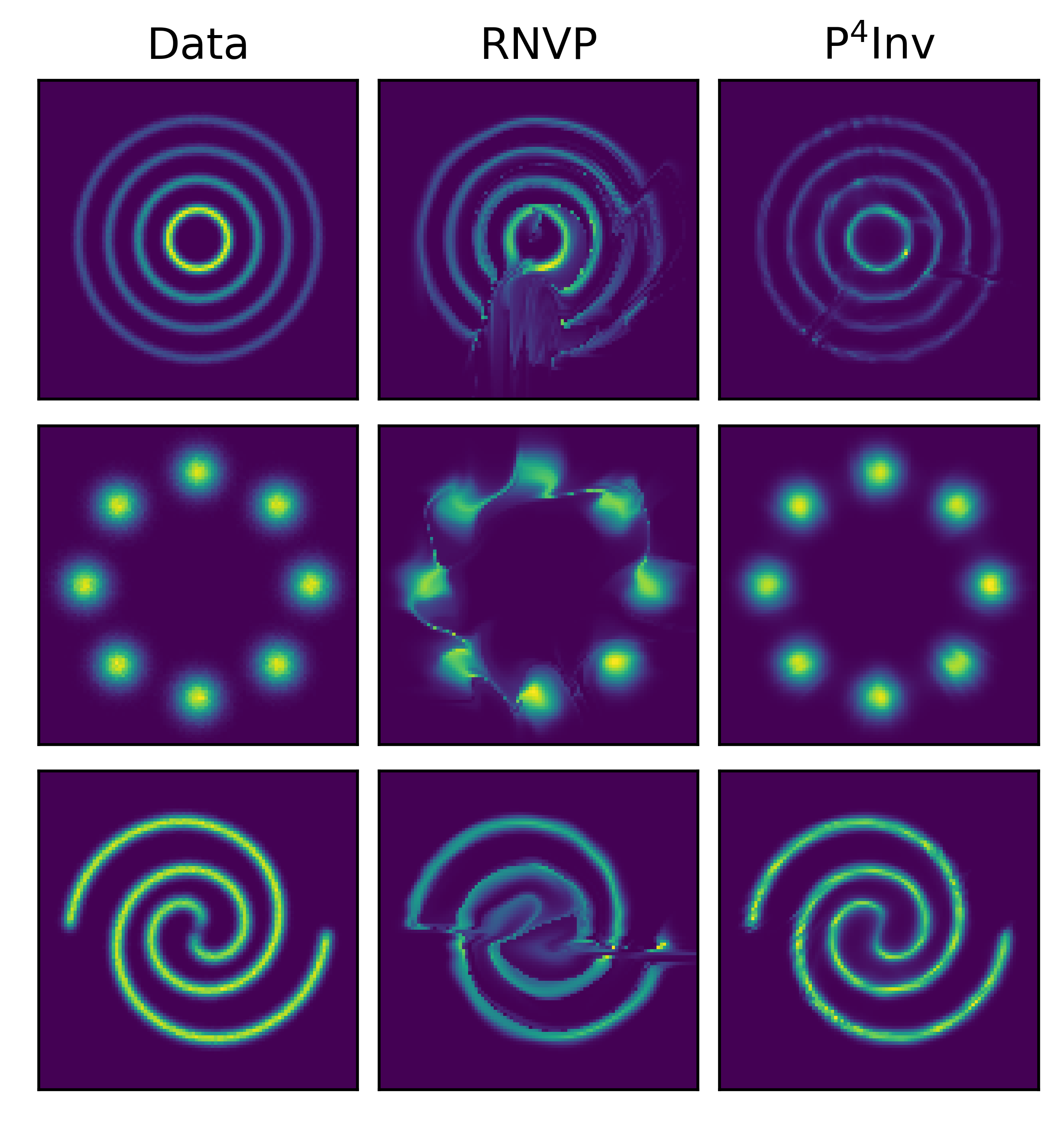}
    \captionof{figure}{
    Density estimation for two-dimensional distributions from RealNVP (RNVP) and P$^4$Inv networks with similar numbers of tunable parameters.
    \label{fig:images}
    }
\end{minipage}%
\end{figure}

Figure \ref{fig:images} compares the generated distributions from the two models. The samples from the P$^4$Inv model aligned favorably with the ground truth. In particular, they reproduced the multimodality of the data. In contrast to RNVP, P$^4$Inv cleanly separated the modes, which underlines the favorable mixing achieved by general linear layers with elementwise nonlinear activations.

\subsection{Boltzmann Generators of Alanine Dipeptide}
Boltzmann generators \citep{noe2019boltzmann} combine normalizing flows with statistical mechanics in order to draw direct samples from a given target density, e.g. given by a many-body physics system. This setup is ideally suited to assess the inversion of normalizing flows since the given physical potential energy defines the target density and thereby provides a quantitative measure for the sample quality. In molecular examples, specifically, the target densities are multimodal, contain singularities, and are highly sensitive to small perturbations in the atomic positions. Therefore, the generation of the 66-dimensional alanine dipeptide conformations is a highly nontrivial test for generative models.

The training efficiency and expressiveness of Boltzmann Generators (see Appendix \ref{sec:bg} for details) were compared between pure RNVP baseline models as used in \citeauthor{noe2019boltzmann} and models augmented by P$^4$Inv swaps (see Section \ref{sec:usage}). The deep neural network architecture and training strategy are described in Appendix \ref{sec:trainbg}. Both flows had 25 blocks as from Figure \ref{fig:bgarchitecture} in the appendix, resulting in 735,050 RNVP parameters. In contrast, the P$^4$Inv blocks had only 9,000 tunable parameters. Due to this discrepancy and the depth of the network, we cannot expect dramatic improvements from adding P$^4$Inv swaps. However, significant numerical errors in the inversion would definitely show in such a setup due to the highly sensitive potential energy.

\begin{figure}[h]
\begin{center}
\includegraphics[width=5cm]{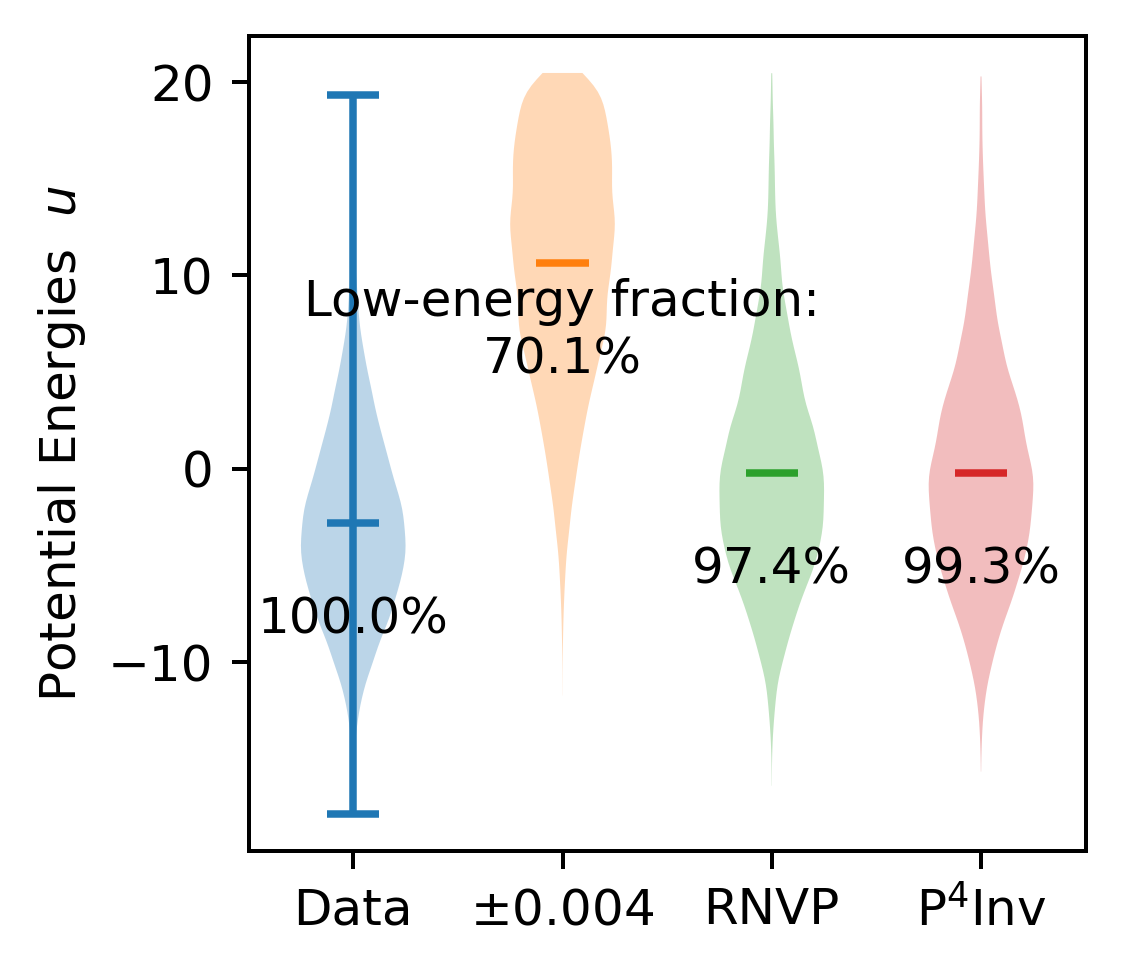}
\includegraphics[width=8.5cm]{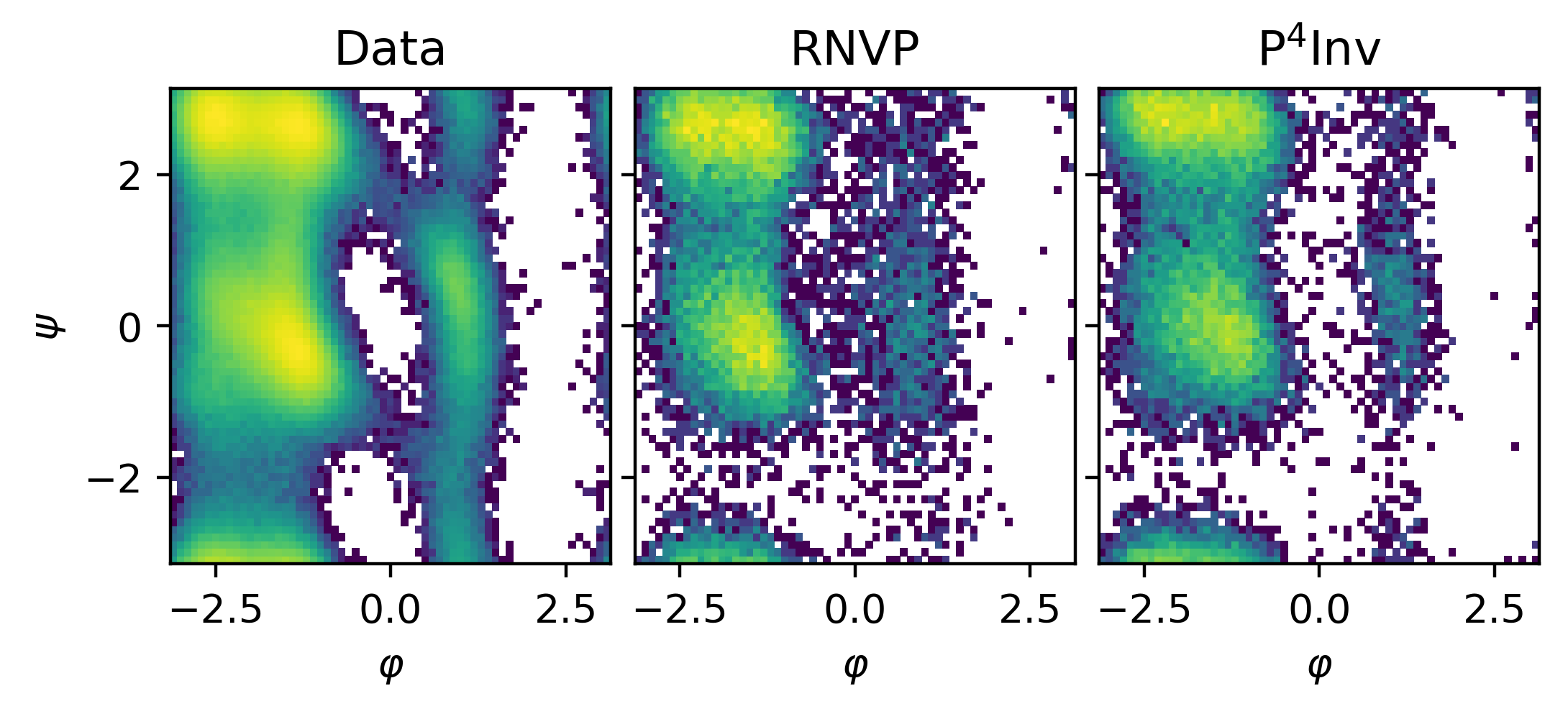}
\end{center}
\caption{
Left: Energy distributions of generated samples in dimensionless units of $k_B T$; the second (orange) violin plot shows energies when the training data was perturbed by normal distributed random noise with 0.004 nm standard deviation. The low-energy fraction for each column denotes the fraction of samples that had potential energy $u$ lower than the maximum energy from the training set ($\approx$ 20 $k_B T$). Right: Joint marginal distribution of the backbone torsions $\varphi$ and $\psi$: training data compared to samples from RealNVP Boltzmann generators with and without P$^4$Inv swaps (denoted \textit{P$^4$Inv} and \textit{RNVP}, respectively).
\label{fig:bgresults}
}
\end{figure}

Figure \ref{fig:bgresults} (left) shows the energy statistics of generated samples. 
To demonstrate the sensitivity of the potential energy, the training data was first perturbed by 0.004 nm  (less than 1\% of the total length of the molecule) and energies were evaluated for the perturbed data set.
As a consequence, the mean of the potential energy distribution increased by 13 $k_B T$. 

In comparison, the Boltzmann generators produced much more accurate samples. The energy distributions from RNVP and P$^4$Inv blocks were only shifted upward by $\approx$ 2.6 $k_B T$ and rarely generated samples with infeasibly large energies. The performance of both models was comparable with slight advantages for models with P$^4$Inv swaps. 
This shows that the P$^4$Inv inverses remained intact during training.
Finally, Figure \ref{fig:bgresults} (right) shows the joint distribution of the two backbone torsions. Both Boltzmann generators reproduced the most important local minima of the potential energy. As in the 2D toy problems, the P$^4$Inv layers provided a cleaner separation of modes.

\section{Other Potential Applications of P$^4$ Updates}
Perturbation theorems are ubiquitous in mathematics and physics so that P$^4$ updates will likely prove useful for retaining other properties of individual layers or neural networks as a whole. To this end, the P$^4$ scheme in Section \ref{sec:general} is formulated in general terms.
Orthogonal matrices may be parameterized in a similar manner to P$^4$Inv through Givens rotations or double-Householder updates. Optimizers that constrain a joint property of multiple layers have previously been used to enforce Lipschitz bounds (\cite{Gouk2018}, \cite{yoshida2017spectral}) and could also benefit from the present work. Applications in physics often rely on networks that obey the relevant physical invariances and equivariances (e.g. \cite{kohler2020equivariant, boyda2020sampling, kanwar2020equivariant, Hermann2020, PhysRevResearch.2.033429, rezende2019equivariant}). These properties might also be amenable to P$^4$ training if suitable property-preserving perturbations can be defined.


\section{Conclusions}
We have introduced P$^4$Inv updates, a novel algorithmic concept to preserve tractable inversion and determinant computation of linear layers using parameterized perturbations. Applications to normalizing flows proved the efficiency and accuracy of the inverses and determinants during training.
A crucial aspect of the $P^4$ method is its decoupled merging step, which allows stable and efficient updates. As a consequence, the invertible linear P$^4$Inv layers can approximate any well-conditioned regular matrix. This feature might open up new avenues to parameterize useful subsets of $\mathrm{GL}(n)$ through penalty functions.
Since perturbation theorems like the rank-one update exist for many classes of linear and nonlinear functions, we believe that the P$^4$ concept presents an efficient and widely applicable way of preserving desirable network properties during training.

\section*{Acknowledgements}
We thank the anonymous reviewers at ICLR 2020 for their valuable suggestions that helped a lot in improving the manuscript. This research was funded through grants by the European Commission (ERC CoG 772230), the German Ministry for Education and Research (Berlin Institute for the Foundations of Learning and Data BIFOLD), the Deutsche Forschungsgemeinschaft (SFB1114/A04 and GRK DAEDALUS), and the Berlin Mathematics research center Math+ (projects AA1-6 and EF2-1).

\bibliography{iclr2021_conference}
\bibliographystyle{iclr2021_conference}

\clearpage

\appendix

\section{Sanity check for the Rank-One Update}
\label{sec:sanity}
Based on the matrix determinant lemma (\eqref{eq:rankone_det}) rank-one updates are ill-conditioned if the term $G:=1+\vv^{T}\mA_\mathrm{inv}\vu$ vanishes. If such a perturbation is ever merged into the network parameters, the stored matrix determinant and inverse degrade and cannot be recovered. 
Therefore, merges are only accepted if the following conditions hold:
\begin{equation*}
    C_{\min}^{(0)}  \leq \ln |G|  \leq C_{\max}^{(0)} \qquad 
    \mathrm{and} \qquad 
    C_{\min}^{(1)}  \leq \ln | G \det \mA|  \leq C_{\max}^{(1)}.
\end{equation*}
The constants $C_{\min}$ and $C_{\max}$ regularize the matrix $\mA$ and its inverse $\mA_\mathrm{inv},$ respectively, since $\ln |\det \mA| = - \ln |\det \mA_\mathrm{inv}|.$

The penalty function is also based on these constraints:
\begin{equation*}
\begin{split}
    g(\vu, \vv) = C_p \cdot \Big( 
        &\mathrm{ReLU}^2 \left( \ln |G| - C_{\max} \right)
         + \mathrm{ReLU}^2 \left( C_{\min} - \ln |G| \right) \\
         + &\mathrm{ReLU}^2 \left( \ln |G \det\mA| - C_{\max} \right)
         + \mathrm{ReLU}^2 \left( C_{\min} - \ln |G \det\mA| \right)
        \Big)
\end{split}
\end{equation*}
with a penalty parameter $C_p.$
For the experiments in this work, we used $C_{\min}=-2$, $C_{\max}=15$, $C_{\min}^{(0)}=-6$, $C_{\max}^{(0)}=\inf$, $C_{\min}^{(1)}=-2.5$, and $C_{\max}^{(1)}=15.5$.

\section{Implementation of P$^4$Inv Layers}

\label{sec:implementation}
In practice, the P$^4$Inv layer stores the current inverse $\mA_\mathrm{inv} \approx \mA^{-1}$ and determinant alongside $\mA.$ The forward pass of an input vector $\vx$ can be computed efficiently by first computing $\vu^T \vx$ and then adding $\vv \vu^T \vx$ to $\mA \vx$. The inverse pass can be similarly structured to avoid any matrix multiplies.

Note that P$^4$ training can straightforwardly be applied to only a part of the network parameters. In this case, all other parameters are directly optimized without the perturbation proxy and the gradient of the loss function $J$ is composed of elements from $\partial J/ \partial \tB$ and $\partial J/ \partial \tA.$
Furthermore, the perturbation of parameters and evaluation of the perturbed model can sometimes be done more efficiently in one step. Also, the merging step from \eqref{eq:merge1} and \eqref{eq:merge2} can additionally be augmented to rectify numerical errors made during optimization. 

In order to allow crossings of otherwise inaccessible regions of the parameter space the merging step was accepted every $N_{\mathrm{force}}$ merges, even if the determinant was poorly conditioned. If $\vu$ or $\vv$ ever contain non-numeric values, merging steps were rejected and the perturbation is reset without performing a merge.

\section{Normalizing Flows}
\label{sec:nf}
A \emph{Normalizing flow} is a learnable bijection $f \colon \mathbb{R}^n \rightarrow \mathbb{R}^n$ which transforms a simple base distribution $p(z)$, by first sampling $z \sim p(z)$, and then transforming into $x = f(z)$. According to the change of variables, the transformed sample $x$ has the density:
\begin{align*}
    q(x) = p\left(f^{-1}(x)\right) \left| \det \mathbf{J}_{f^{-1}}(x) \right|.
\end{align*}
Given a target distribution $\rho(x)$, this tractable density allows minimizing the reverse Kullback-Leibler (KL) divergence
$
    \mathcal{D}_{\text{KL}}\left[ q(x) \| \rho(x) \right],
$
e.g., if $\rho(x)$ is known up to a normalizing constant, or the forward KL divergence 
$
    \mathcal{D}_{\text{KL}}\left[ \rho(x) \| q(x) \right],
$
if having access to samples from $\rho(x)$.

\section{Coupling Layers}
Maintaining invertibility and computing the inverse and its Jacobian is a challenge, if $f$ could be an arbitrary function. Thus, it is common to decompose $f$ in a sequence of \textit{coupling layers}
\begin{align*}
    f = g^{(1)} \circ \mathcal{S}^{(1)} \circ  \ldots g^{(K)} \circ \mathcal{S}^{(K)}.
\end{align*}
Each $g^{(k)}$
is constrained to the form 
$g^{(k)}(x) = T^{(k)}(x_{1}, x_{2}) \oplus x_{2}$, where $x = x_1 \oplus x_2,~  x_{1} \in \mathbb{R}^{m}$ and $x_{2} \in \mathbb{R}^{n-m}$. Here $T^{(k)} \colon \mathbb{R}^{m} \times \mathbb{R}^{n-m} \rightarrow \mathbb{R}^{m}$ is a point-wise transformation, which is invertible in its first component given a fixed $x_{2}$. Possible implementations include simple affine transformations \citep{dinh2014nice,dinh2016density} as well as complex transformations based on splines \citep{muller2018neural, durkan2019neural}. Each $g^{(k)}$ thus has a block-triangular Jacobian matrix
\begin{align*}
    \mathbf{J}_{g^{(k)}} = \left[ \begin{array}{cc}
        \mathbf{J}_{T^{(k)}} &  \mathbf{M}^{(k)} \\
        \mathbf{0} & \mathbf{I}_{m \times m}
    \end{array} \right],
\end{align*}
where $\mathbf{J}_{T^{(k)}}$ is a $(n-m) \times (n-m)$ diagonal matrix. The layers $\mathcal{S}^{(k)}$ take care of achieving a mixing between coordinates and are usually represented as simple swaps
\begin{align*}
    \mathcal{S}^{(k)} = \left[ \begin{array}{cc}
        \mathbf{0} &  \mathbf{I}_{n-m \times n-m} \\
        \mathbf{I}_{m \times m} & \mathbf{0}
    \end{array} \right].
\end{align*}
The total log Jacobian of $f_{\theta}$ is then given by
\begin{align*}
    \log \det \mathbf{J}_{f_{\theta}} = \sum_{k=1}^{K} \text{tr}\left[\log\left( \mathbf{J}_{T^{(k)}}\right)\right] + \log\det \mathcal{S}^{(k)},
\end{align*}
where $\log\det \mathcal{S}^{(k)} = 0$ when $\mathcal{S}^{(k)}$ is given by the simple swaps above.

\section{Boltzmann Generators}
\label{sec:bg}
Boltzmann Generators \citep{noe2019boltzmann} are generative neural networks that sample with respect to a known target energy, as for example given by molecular force fields. The potential energy $u:\mathbb{R}^{3n} \rightarrow {\mathbb{R}}$ of such systems is a function of the atom positions $\rvx.$ The corresponding probability of each configuration $\rvx$ is given by the Boltzmann distribution $p_x(\rvx) = \exp(-\beta u(\rvx))/Z,$ where $\beta=1/(k_B T)$ is inverse temperature with the Boltzmann constant $k_B.$ The normalization constant $Z$ is generally not known.

The generation uses a normalizing flow and training is performed via a combination of density estimation and energy-based training.
Concretely, the following loss function is minimized
\begin{equation} 
\label{eq:loss}
J(\mA) = w_{l} J_{l}(\mA) + w_{e} J_{e}(\mA),
\end{equation}
where $w_{l} + w_{e}=1$ denote weighting factors between density estimation and energy-based training.
The maximum likelihood and KL divergence in \eqref{eq:loss} are defined respectively as 
\begin{align*}
    J_l &= \mathbb{E}_{\rvx \sim p_x}\left[ \frac{1}{2} \| F_{xz}(\rvx; \mA) \|^2 - \ln R_{xz}(\rvx; \mA)  \right] \quad \mathrm{and} \\ 
    J_e &= \mathbb{E}_{\rvz \sim p_z} \left[ u(F_{zx}(\rvz; \mA)) - \ln R_{zx}(\rvz; \mA) \right].
\end{align*}
  
As an example, we train a model for alanine dipeptide, a molecule with 22 atoms, in water. Water is represented by an implicit solvent model. This system was previously used in \cite{wu2020stochastic}. Training data was generated using MD simulations at 600K to sample all metastable regions.

\section{Training of Linear Toy Problems}
\label{sec:traintoy}

The P$^4$Inv layers were trained using a stochastic gradient descent optimizer with a learning rate of $10^{-2}$ and the hyperparameters from Table \ref{tab:toyparams}. The matrices were initialized with the identity.

\begin{table}[h!]
\caption{Hyperparameters for linear toy problems
\label{tab:toyparams}
}
\begin{center}
\begin{tabular}{lcccc}
Problem (Dimension) & $N$ & $N_\mathrm{force}$ & $N_\mathrm{correct}$ & Penalty $C_0$
\\ \hline \\
Positive Definite  (32) & 1, 10, 50 & 10 & 50 & 0.1 \\
Orthogonal (128)     &  1   & 10 & 50 & 0.1 \\
$-\mI$ (101)       & 1 & 10 & 50 & 0.0 \\
\end{tabular}
\end{center}
\end{table}

\section{Training of 2D Distributions}
\label{sec:train2D}
The P$^4$Inv layers used for 2D distributions were composed of blocks containing
\begin{enumerate}
\item a P$^4$Inv layer with bias (2D),
\item an elementwise Bent identity,
\item another P$^4$Inv layer with bias (2D), and
\item an inverse elementwise Bent identity.
\end{enumerate}
100 of these blocks were stacked resulting in 1200 tunable parameters (counting elements of $\vu$ and $\vv$). P$^4$Inv training was performed with $N=10,$ $N_\mathrm{force}=10$ and $N_\mathrm{correct}=50$. No penalty was used. Matrices were initialized with the identity $\mI_2.$

The RealNVP network used for comparison was composed of five RealNVP layers. The additive and multiplicative conditioner networks used dense nets with two 6-dimensional hidden layers each and $\tanh$ activation functions, respectively. This resulted in a total of 1230 parameters.

The examples are taken from \cite{grathwohl2018ffjord}. Priors were two-dimensional standard normal distributions. Adam optimization was performed for 8 epochs of 20000 steps and with a batch size of 200. The initial learning rate was $5\cdot 10^{-3}$ and decreased by a factor of $0.5$ in each epoch. After each merging step, the metaparameters of the Adam optimizer were reset to their initial state.

\begin{figure}[h!]
    \centering
    \includegraphics[width=0.6\linewidth]{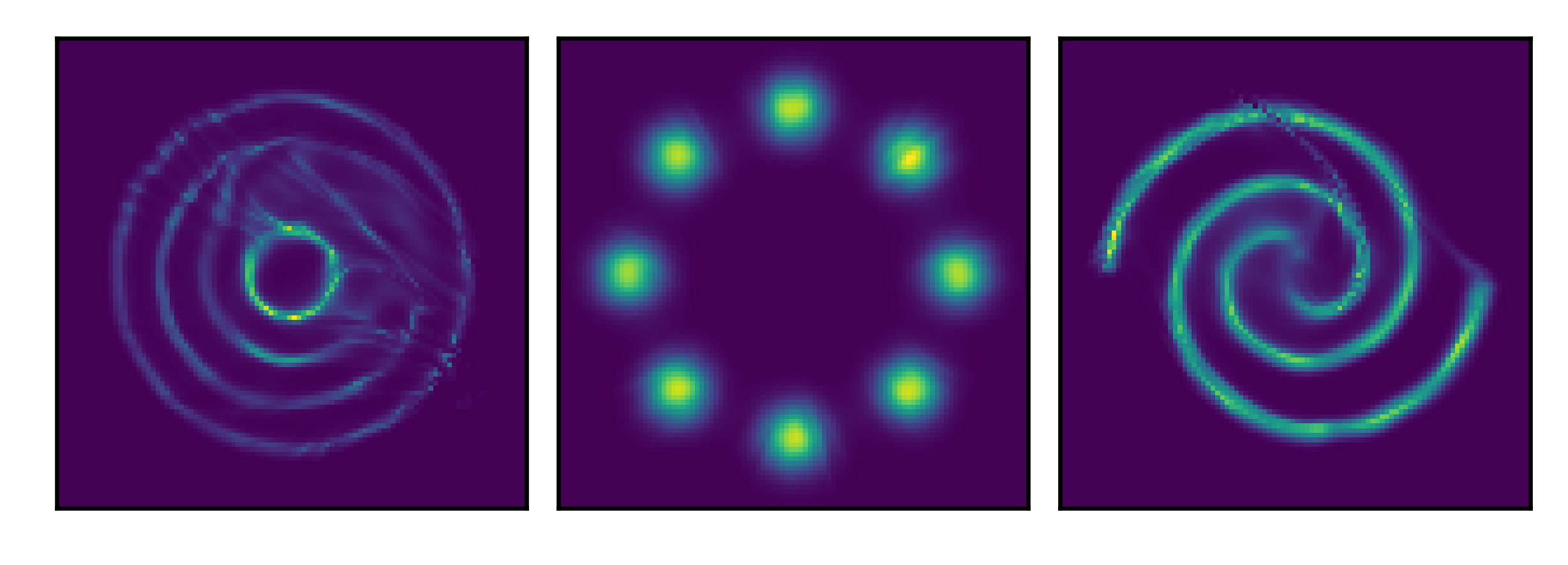}
    \caption{Samples from P$^4$Inv training via the inverse.}
    \label{fig:ibent}
\end{figure}

Figure \ref{fig:ibent} complements Figure \ref{fig:images} by showing samples from a network with only inverse P$^4$Inv blocks. While the samples are worse than with forward blocks, the distributions are still well represented. This result indicates that the premature convergence encountered for linear test problems is a lesser problem in nonlinear problems and deep architectures.

\section{Training of Boltzmann Generators}
\label{sec:trainbg}

\begin{figure}[h]
\begin{center}
\includegraphics[width=7cm]{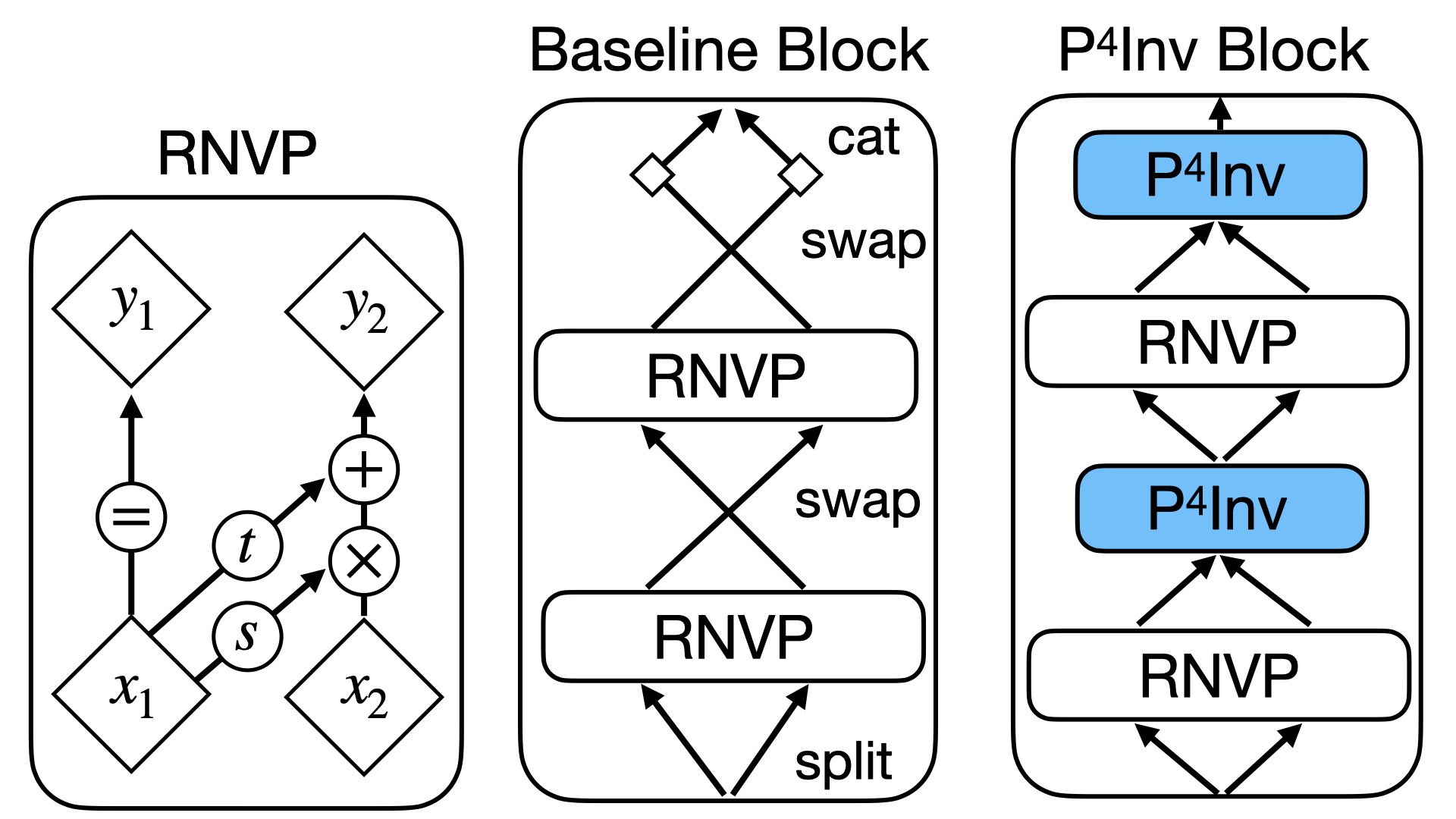}
\end{center}
\caption{Neural network blocks used in the Boltzmann generator application. The baseline architecture is a normalizing flow composed of Real NVP (RNVP) coupling blocks. RNVP uses input from two channels $x_1$ and $x_2.$ The input from the first channel is left untouched, $y_1=x_1,$ while the output $y_2$ from the second channel is conditioned on the first channel through two neural networks $t$ and $s$. Each block of the baseline model contains two RNVP blocks and two swapping steps that are bracketed by splitting and concatenation (\texttt{cat}) of the data. Instead of the swapping steps, the P$^4$Inv model uses invertible linear layers that are trained through P$^4$ updates.
\label{fig:bgarchitecture}
}
\end{figure}

The normalizing flows in Boltzmann generators were composed of the blocks shown in Figure \ref{fig:bgarchitecture} and a mixed coordinate transform as defined in \cite{noe2019boltzmann}. The test problem was taken from \cite{dibak2020neurips}.
RNVP layers contained two 60-dimensional hidden layers each and ReLU and $\tanh$ activation functions for both $t$ and $s$, respectively. The baseline model consisted of blocks of alternated RNVP blocks and swaps. The P$^4$Inv model used invertible linear layers instead of the swapping of input channels in the baseline model. The computational overhead due to this change was negligible. RNVP parameters were optimized directly as usual and only the P$^4$Inv layers are affected by the P$^4$ updates.
Merging was performed every $N=100$ steps with $N_\mathrm{force}=10$ and $N_\mathrm{correct}=50$. No penalty was used, i.e. $C_0=0.0$. The P$^4$Inv matrices were initialized with the reverse permutation, i.e. $\emA_{ij}=\delta_{i(n-j)}.$

Density estimation with Adam was performed for 40,000 optimization steps with a batch size of 256 and a learning rate of $10^{-3}.$ A short energy-based training epoch was appended for 2000 steps with a learning rate of $10^{-5}$ and $w_e/w_l=0.05.$ After each merging step, the metaparameters of the Adam optimizer were reset to their initial state for all P$^4$Inv parameters.

\section{MNIST Classification via Rank-one Updates}
\label{sec:mnist}
Compared to fully-connected multi-layer perceptrons (MLP), rank-one updates reduce the number of independent trainable parameters per layer from $m\cdot n$ to $m + n - 1,$ where $m$ and $n$ are the input and output dimension, respectively. It is therefore useful to study how the reduced search space dimension affects the training efficiency in a nonlinear setup. To this end, non-invertible classifier MLPs were trained on MNIST with an unrestricted search space as well as through rank-one updates. The original network was composed of two convolutional layers, two dropout layers, and two linear layers. In P$^4$ training, the two linear layers (dimensions 9216$\times$128 and 128$\times$10) were trained through rank-one updates which were merged in every iteration ($N=1$). Naturally, this training did not involve any inverse or determinant updates. A vanilla SGD optimizer was used with various learning rates.

\begin{figure}[tbhp]
\begin{center}
\includegraphics[width=\linewidth]{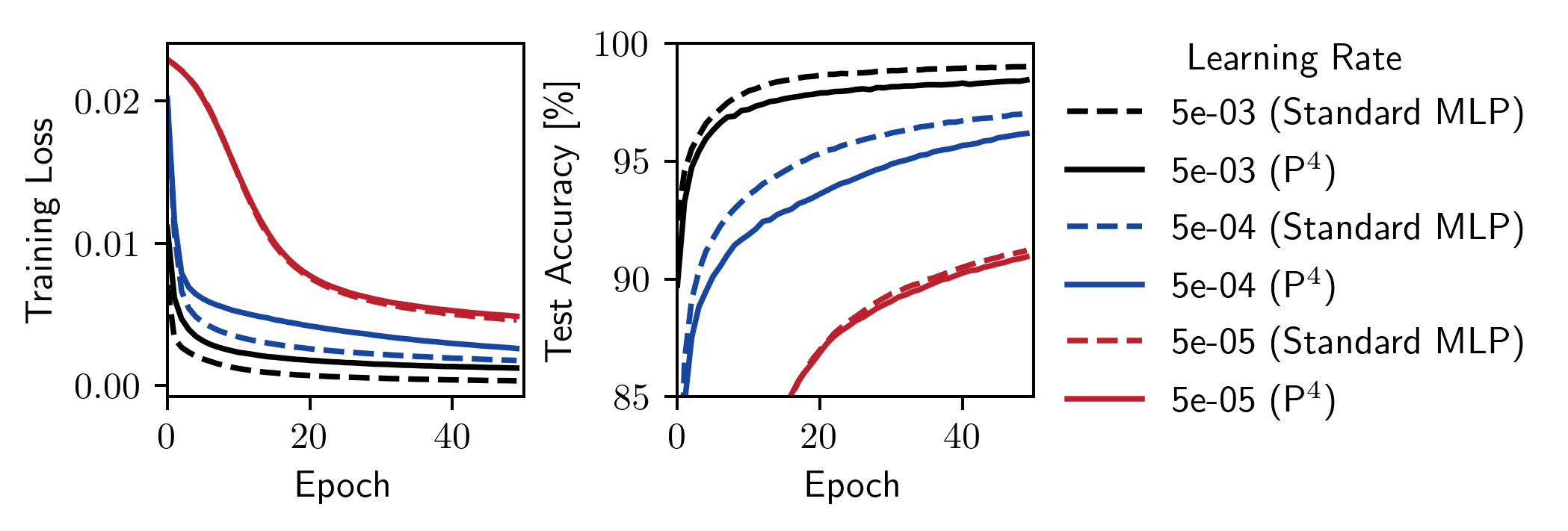}
\end{center}
\caption{
Training loss and test accuracy during MNIST training with a vanilla SGD optimizer averaged over ten replicas. Standard multilayer perceptrons (MLP) are compared with rank-one updates.
\label{fig:mnist}
}
\end{figure}

Figure \ref{fig:mnist} shows the training loss and test accuracy during training. As for the linear problems from the previous subsection, the training efficiency was virtually unaffected during the first phase of the optimization, i.e. when the descent direction did not change significantly between subsequent iterations. However, as the descent direction became more noisy in the vicinity of the optimum, the training with rank-one updates became less efficient.

\end{document}